\documentclass[runningheads]{llncs}
\usepackage{graphicx}
\usepackage{comment}
\usepackage{amsmath,amssymb} 
\usepackage{color}

\usepackage[width=122mm,left=12mm,paperwidth=146mm,height=193mm,top=12mm,paperheight=217mm]{geometry}

\usepackage{array}
\usepackage{amsmath}
\usepackage{url}
\usepackage{bm}
\usepackage[]{hyperref}

\begin{document}
\pagestyle{headings}
\mainmatter
\def\ECCVSubNumber{4582}  

\title{Adaptive Mixture Regression Network with Local Counting Map
	for Crowd Counting} 



\titlerunning{ }
%

\author{
Xiyang Liu \inst{1}  \quad
Jie Yang\inst{2} \quad
Wenrui Ding \inst{1\thanks{Corresponding author}  }
}

\authorrunning{ }
%
\institute{
Beihang University, China  \and 
Shunfeng Technology (Beijing) Co., Ltd   \\
}

\maketitle

\vspace{-18pt}
\begin{abstract}
The crowd counting task aims at estimating the number of people located in an image or a frame from videos.
Existing methods widely adopt density maps as the training targets to optimize the point-to-point loss.
While in testing phase, we only focus on the differences between the crowd numbers and the global summation of density maps, which indicate the inconsistency between the training targets and the evaluation criteria.
To solve this problem, we introduce a new target, named local counting map (LCM), to obtain more accurate results than density map based approaches.
Moreover, we also propose an adaptive mixture regression framework with three modules in a coarse-to-fine manner to further improve the precision of the crowd estimation: scale-aware module (SAM), mixture regression module (MRM) and adaptive soft interval module (ASIM).
Specifically, SAM fully utilizes the context and multi-scale information from different convolutional features;
MRM and ASIM perform more precise counting regression on local patches of images.
Compared with current methods, the proposed method reports better performances on the typical datasets.
The source code is available at \url{https://github.com/xiyang1012/Local-Crowd-Counting} .


\end{abstract}
\vspace{-20pt}

\section{Introduction}
\vspace{-10pt}

The main purpose of visual crowd counting is to estimate the numbers of people from static images or frames.
Different from pedestrian detection \cite{li2017scale,mao2017can,liu2019high}, crowd counting datasets only provide the center points of heads, instead of the precise bounding boxes of bodies.
So most of the existing methods draw the density map \cite{NIPS2010_4043} to calculate crowd number.
For example, CSRNet \cite{CSRNet_2018_CVPR} learned a powerful convolutional neural network (CNN) to get the density map with the same size as the input image.
Generally, for an input image, the ground truth of its density map is constructed via a Gaussian convolution with a fixed or adaptive kernel on the center points of heads .
Finally, the counting result can be represented via the summation of the density map.

In recent years, benefit from the powerful representation learning ability of deep learning, crowd counting researches mainly focus on CNN based methods \cite{MCNN_2016_CVPR,CPCNN_2017_ICCV,SANet_2018_ECCV,sam2018top,babu2018divide} to generate high-quality density maps.
The mean absolute error (MAE) and mean squared error (MSE) are adopted as the evaluation metrics of crowd counting task.
However, we observed an inconsistency problem for the density map based methods: the training process minimizes the $L_1/L_2$ error of the density map, which actually represents a point-to-point loss \cite{SPANet_2019_ICCV}, while the evaluation metrics in the testing stage only focus on the differences between the ground-truth crowd numbers and the overall summation of the density maps.
Therefore, the model with minimum training error of the density map does not ensure the optimal counting result when testing.

\begin{figure}[t]
	\centering
	\scalebox{0.35} {\includegraphics{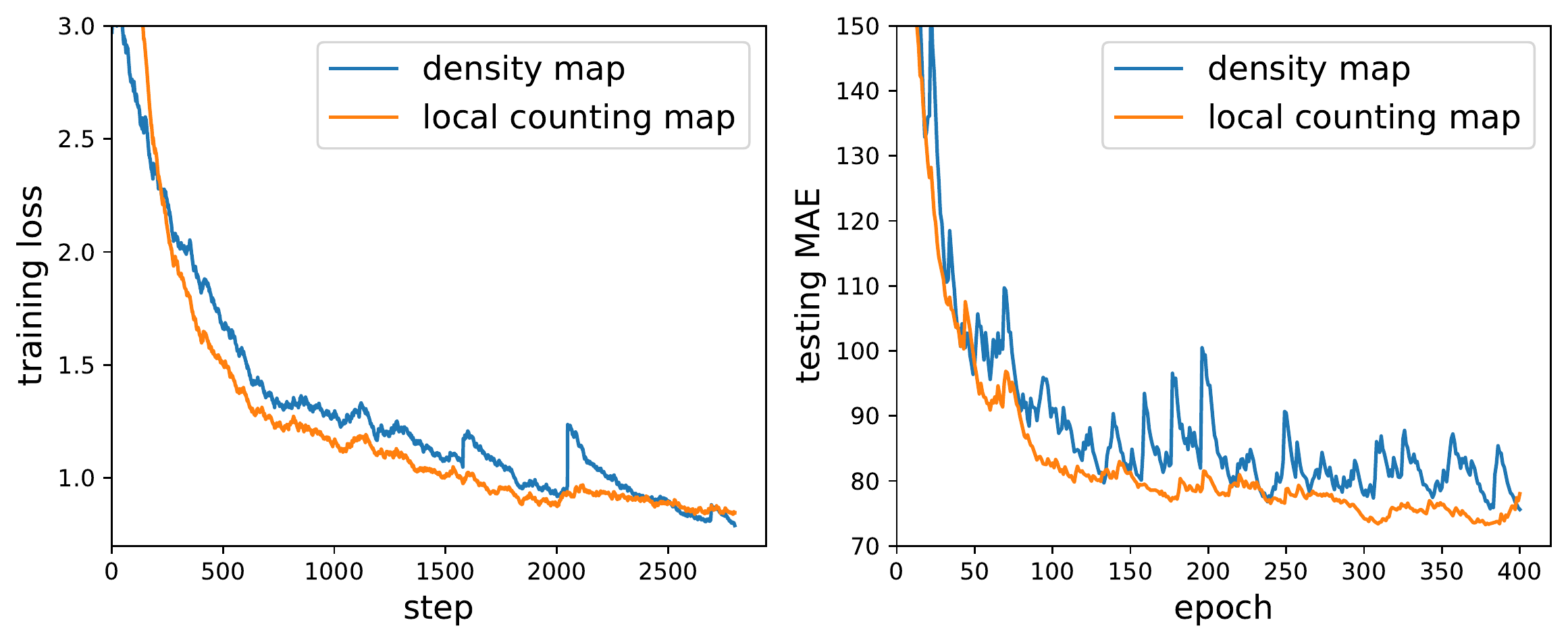}}
	\caption{
		Training loss curves (\textit{left}) and testing loss curves (\textit{right}) between 
		the two networks sharing VGG16 as the backbone with different regression targets, 
		density map and local counting map on ShanghaiTech Part A dataset.
		The network trained with the local counting map has the lower error and more stable performance on the testing dataset than the one with the density map
	}		
	\label{fig1}
\end{figure}

\begin{figure}[t]
	\centering
	\scalebox{0.31} {\includegraphics{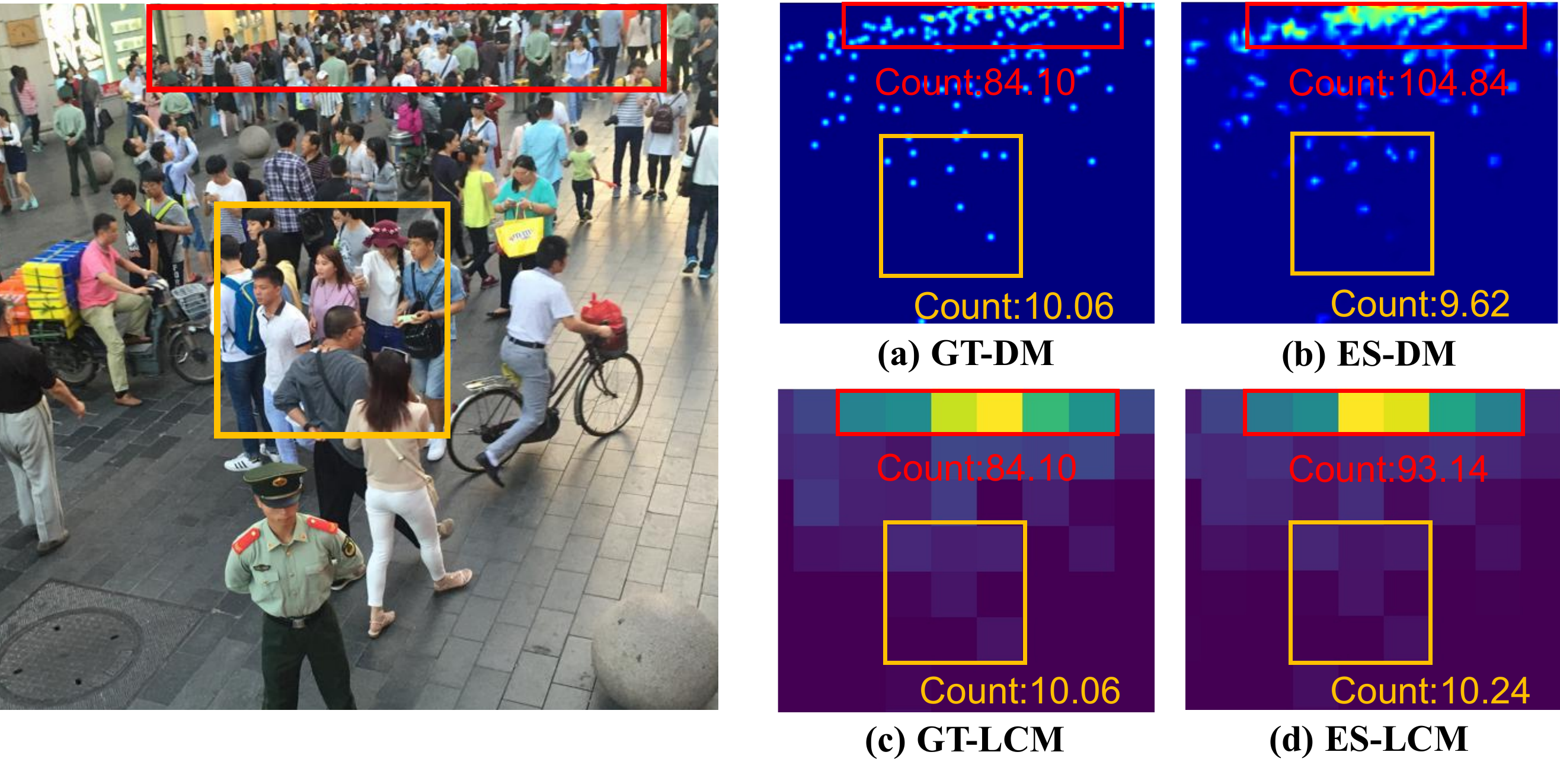}}
	\caption{An intuitive comparison between the local counting map (LCM) and the density map (DM) on local areas.
	LCM has more accurate estimation counts on both dense (the red box) and sparse (the yellow box) populated areas.
		(\textit{GT-DM}: ground-truth of DM;   \textit{ES-DM}: estimation of DM;
		\textit{GT-LCM}: ground-truth of  LCM;    \textit{ES-LCM}: estimation of LCM)
	}
	\label{fig2}
\end{figure}

To draw this issue, we introduce a new learning target, named local counting map (LCM), in which each value represents the crowd number of a local patch rather than the probability value indicating whether has a person or not in the density map.
In Sec. \ref{Sec 3.1}, we prove that LCM is closer to the evaluation metric than the density map through a mathematical inequality deduction.
As shown in Fig. \ref{fig1}, LCM markedly alleviates the inconsistency problem brought by the density map.
We also give an intuitive example to illustrate the prediction differences of LCM and density map.
As shown in Fig. \ref{fig2}, the red box represents the dense region and the yellow one represents the sparse region.
The prediction of density map is not reliable in dense areas,
while LCM has more accurate counting results in these regions.

To further improve the counting performance, we propose an adaptive mixture regression framework to give an accurate estimation of crowd numbers in a coarse-to-finer manner.
Specifically, our approach mainly includes three modules:
1) scale-aware module (SAM) to fully utilize the context and multi-scale information contained in feature maps from different layers for estimation;
2) mixture regression module (MRM) and 
3) adaptive soft interval module (ASIM) to perform precise counting regression on local patches of images.

In summary, the main contributions in this work are in the followings:
\begin{itemize}
	\item  We introduce a new learning target LCM, which alleviates the inconsistency problem between training targets and evaluation criteria, and reports better counting performance compared with the density map.
	\item We propose an adaptive mixture regression framework in a coarse-to-finer manner, which fully utilizes the context and multi-scale information from different convolutional features and performs more accurate counting regression on local patches.
\end{itemize}

The rest of the paper is described as follows:
Sec. \ref{Sec 2} reviews the previous work of crowd counting;
Sec. \ref{Sec 3} details our method;
Sec. \ref{Sec 4} presents the experimental results on typical datasets;
Sec. \ref{Sec 5} concludes the paper.

\section{Related Work}{\label{Sec 2}}

Recently, CNN based approaches have become the focus of crowd counting researches.
According to regression targets, they can be classified into two categories: density estimation based approaches and direct counting regression ones.

\subsection{Density Estimation based Approaches}

The early work \cite{NIPS2010_4043} defined the concept of density map and transformed the counting task to estimate the density map of an image. The integral of density map in any image area is equal to the count of people in the area.
Afterwards, Zhang \textit{et al.} \cite{Zhang_2015_CVPR} used CNN to regress both the density map and the global count. It laid the foundation for subsequent works based on CNN methods.
To improve performance, some methods aimed at improving network structures.
MCNN \cite{MCNN_2016_CVPR} and Switch-CNN \cite{SwitchCNN_2017_CVPR} adopted multi-column CNN structures for mapping an image to its density map.
CSRNet \cite{CSRNet_2018_CVPR} removed multi-column CNN and used dilated convolution to expand the receptive field.
SANet \cite{SANet_2018_ECCV} introduced a novel encoder-decoder network to generate high-resolution density maps.
HACNN \cite{HACNN_2019_TIP} employed attention mechanisms at various CNN layers to selectively enhance the features.
PaDNet \cite{tian2019padnet} proposed a novel end-to-end architecture for pan-density crowd counting.
Other methods aimed at optimizing the loss function.
ADMG \cite{adaptive_2019_ICCV} produced a learnable density map representation.
SPANet \cite{SPANet_2019_ICCV} put forward MEP loss to find the pixel-level subregion with high discrepancy to the ground truth.
Bayesian Loss \cite{BL_2019_ICCV} presented a Bayesian loss to adopt a more reliable supervision on the count expectation at each annotated point. 

\subsection{Direct Counting Regression Approaches}

Counting regression approaches directly estimate the global or local counting number of an input image.
This idea was first adopted in \cite{chen_2012_BMVC}, which proposed a multi-output  regressor to estimate the counts of people in spatially local regions for crowd counting.
Afterwards, Shang \textit{et al.} \cite{Shang_2016_ICIP} used recurrent network layers to predicted both global and local counts.
Lu \textit{et al.} \cite{lu2017tasselnet} and Paul \textit{et al.} \cite{paul_2017_count}  used overlapping sliding windows to sample local patches intensively and regressed local counts. The inferred redundant local counts were normalized and eventually integrated into the global count.
Chattopadhyay \textit{et al.} \cite{chattopadhyay2017counting} employed a divide and conquer strategy while incorporating context across the scene to adapt the subitizing idea to counting.
Stahl \textit{et al.} \cite{stahl2018divide} adopted a local image divisions method to predict global image-level counts without using any form of local annotations.
S-DCNet \cite{S-DCNet_2019_ICCV} exploited a spatial divide and conquer network that learned from closed set and generalize to open set scenarios.

Though many approaches have been proposed to generate high-resolution density maps or predict global and local counts, the robust crowd counting of diverse scenes remains hard.
Different with previous methods, we firstly introduce a novel regression target, and then adopt an adaptive mixture regression network in a coarse-to-fine manner for better crowd counting.

\section{Proposed Method}{\label{Sec 3}}

\begin{figure}[t]
	\centering
	\scalebox{0.28} {\includegraphics{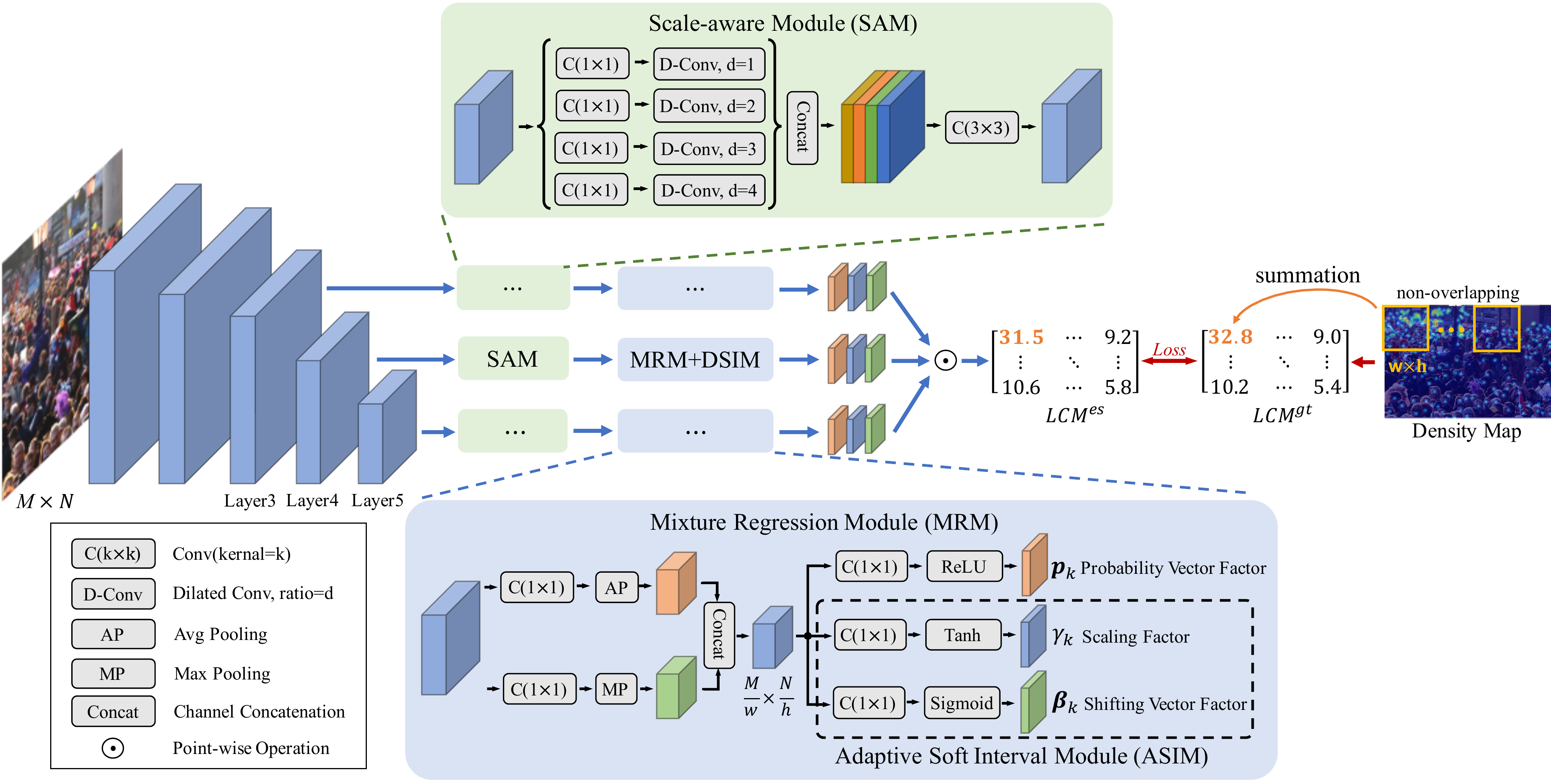}}
	\caption{
		The overview of our framework mainly including three modules:
		1) scale-aware module (SAM), used to enhance multi-scale information of feature maps via multi-column dilated convolution;
		2) mixture regression module (MRM) and 3) adaptive soft interval module (ASIM), used to regress feature maps to the probability vector factor $\bm{p}_k$, the scaling factor $\gamma_k$  and the shifting vector factors $\bm{\beta}_k$ of the $k$-th mixture, respectively.
		We adopt the feature maps of layers 3, 4 and 5 as the inputs of SAM.
		The local counting map (LCM) is calculated according to parameters $\{\bm{p}_k, \gamma_k, \bm{\beta}_k \}$ and point-wise operation in Eq. (\ref{regression}).
		For an input $M \!\times\! N$ image and the $w \!\times\! h $ patch size, the output of the entire framework is a $\frac{M}{w} \!\times\! \frac{N}{h}$ LCM
	}
	\label{fig3}	
\end{figure}

In this section, we first introduce LCM in details and prove its superiority compared with the density map in Sec. \ref{Sec 3.1}.
After that, we describe SAM, MRM and ASIM of the adaptive mixture regression framework in Sec. \ref{Sec 3.2}, \ref{Sec 3.3} and \ref{Sec 3.4}, respectively.
The overview of our framework is shown in Fig. \ref{fig3}.

\subsection{Local Counting Map}{\label{Sec 3.1}}

For a given image containing $n$ heads, the ground-truth annotation can be described as $GT(p) = \sum_{i=1}^{n} \delta(p-p_i)$, where $p_i$ is the pixel position of the $i$-th head's center point.
Generally, the generation of the density map is based on a fixed or adaptive Gaussian kernel $G_\sigma$, which is described as $D(p)=\sum_{i=1}^{n} \delta(p-p_i) * G_\sigma$. In this work, we fix the spread parameter $\sigma$ of the Gaussian kernel as $15$.

Each value in LCM represents the crowd number of a local patch, rather than a probability value indicating whether has a person or not in the density map.
Because heads may be at the boundary of two patches in the process of regionalizing an image, it's unreasonable to divide people directly.
Therefore, we generate LCM by summing the density map patch-by-patch.
Then, the crowd number of local patch in the ground-truth LCM is not discrete value, but continuous value calculated based on the density map.
The LCM can be described as the result of the non-overlapping sliding convolution operation as follows:

{\setlength\abovedisplayskip{-3mm}
\setlength\belowdisplayskip{1mm}
\begin{align}
\mathit{LCM} = D * \bm{1}_{(w,h)} ,
\end{align}
}where $D$ is the density map, $\bm{1}_{(w,h)}$ is the matrix of ones and $(w,h)$ is the local patch size.

Next, we explain the reason that LCM can alleviate the inconsistency problem of the density map mathematically.
For a test image, we set the $i$-th pixel in ground-truth density map as $g_i$ and the $i$-th pixel in estimated density map as $e_i$. The total pixels number of the image is $m$ and the pixels number of the local patch is $t=w \times h$.
The evaluation criteria of mean absolute error (MAE), the error of LCM (LCME) and the error of density map (DME) can be calculated as follows:

{\setlength\abovedisplayskip{-3mm}
\setlength\belowdisplayskip{-3mm}
\begin{align}
\mathit{MAE} = &\big| (e_1+e_2+...+e_m) - (g_1+g_2+...+g_m) \big|, \\
\mathit{LCME} = &\big|(e_1+...+e_t) - (g_1+...+g_t)\big| +...+ \nonumber \\
&\big|(e_{m-t}+...+e_m) - (g_{m-t}+...+g_m)\big|, \\
\mathit{DME}  =  &\big| e_1-g_1 \big|+\big| e_2-g_2 \big|+...+ \big|e_m-g_m \big|.
\end{align}
}

According to absolute inequality theorem, we can get the relationship among them:

{\setlength\abovedisplayskip{-3mm}
\setlength\belowdisplayskip{2mm}
\begin{align}
\mathit{MAE}  \leq  \mathit{LCME}  \leq  \mathit{DME} .
\end{align}
}When $t\!=\!1$, we have LCME $\!=\!$ DME. When $t\!=\!m$, we get LCME  $\!=\!$  MAE.
LCME provides a general form of loss function adopted for crowding counting.
No matter what value $t$ takes, LCME proves to be a closer bound of MAE than DME theoretically.


On the other side, we clarify the advantages of LCME for training, compared with DME and MAE.
1) DME mainly trains the model to generate probability responses pixel-by-pixel.
However, pixel-level position labels generated by a Gaussian kernel may be low-quality and inaccurate for training, due to severe occlusions, large variations of head size, shape and density, etc.
There is also a gap between the training loss DME and the evaluation criteria MAE.
So the model with minimum training DME does not ensure the optimal counting result when testing with MAE.
2) MAE means direct global counting from an entire image.
But global counting is an open-set problem and the crowd number ranges from 0 to $\infty$,
the MAE optimization makes the regression range greatly uncertain.
Meanwhile, global counting would ignore all spatial annotated information, which couldn't provide
visual density presentations of the  prediction results.
3) LCM provides a more reliable training label than the density map, which discards the inaccurate pixel-level position information of density maps and focuses on the count values of local patches.
LCME also lessens the gap between DME and MAE.
Therefore, we adopt LCME as the training loss rather than MAE or DME.


\subsection{Scale-aware Module}{\label{Sec 3.2}}

Due to the various shooting views of cameras and the complex distributions of crowd positions, the scales of heads in an image are usually very polytropic, which brings great challenge to crowd counting task. To deal with this problem, we propose scale-aware module (SAM) to enhance the multi-scale feature extraction capability of the network. 
The previous works, such as L2SM \cite{xu2019learn} and S-DCNet \cite{S-DCNet_2019_ICCV}, mainly focused on the fusion of feature maps from different CNN layers and acquire multi-scale information through feature pyramid network structure.
Different from them, the proposed SAM achieves multi-scale information enhancement only on a single layer feature map and performs this operation at different convolutional layers to bring rich information to subsequent regression modules.

For fair comparisons, we treat VGG16 as the backbone network for CNN-based feature extraction. 
As shown in Fig. \ref{fig3}, we enhance the feature maps of layers 3, 4 and 5 of the backbone through SAM, respectively.
SAM first compresses the channel of feature map via $1 \!\times\! 1$ convolution.
Afterwards, the compressed feature map is processed through dilated convolution with different expansion ratios of 1, 2, 3 and 4 to perceive multi-scale features of heads.
The extracted multi-scale feature maps are fused via channel-wise concatenation operation and $ 3\!\times\! 3$ convolution. The size of final feature map is consistent with the input one.

\subsection{Mixture Regression Module}{\label{Sec 3.3}}

Previously, TasselNet \cite{lu2017tasselnet} directly regressed the local features to get the count value of the patch.
However, the counting result obtained in this way is not robust enough, and it is not easy to perform network training and parameters convergence. MRM we proposed can achieve better performance via a coarse-to-fine manner.

First, we discuss the case of coarse regression.
For a certain local patch, we assume that the patch contains the upper limit of the crowd as $C_m$. Thus, the number of people in this patch is considered to be $[0, C_m]$.
We equally divide $[0, C_m]$ into $s$ intervals and the length of each interval is $\frac{C_m}{s}$. The vector $\bm{p}\!=\![p_1, p_2,...,p_s]^T$ represents the probability of $s$ intervals, and the vector $\bm{v} \!=\! [v_1,v_2,...,v_s]^T \!=\! [ \frac{1 \cdot C_m}{s},  \frac{2 \cdot C_m}{s},..., C_m]^T$ represents the value of $s$ intervals.
Then the counting number $C_p$ of a local patch in coarse regression can be obtained as followed:

{\setlength\abovedisplayskip{-2mm}
\setlength\belowdisplayskip{0mm}
\begin{align}
C_p &= \bm{p}^{T}  \bm{v} = \sum_{i=1}^{s} p_i \cdot v_i
= \sum_{i=1}^{s} p_i \cdot \frac{i \cdot C_m}{s} \\
&= C_m \sum_{i=1}^{s} \frac{p_i \cdot i}{s}.
\end{align}
}

Next, we discuss the situation of fine mixture regression.
We assume that the fine regression is consisted of $K$ mixtures.
Then, the interval number of the $k$-th mixture is $s_k$.
The vector $\bm{p}$ of the $k$-th mixture is $\bm{p_k}\!=\![p_{k,1}, p_{k,2},...,p_{k,s}]^T$
and the vector $\bm{v}$ is $\bm{v_k} \!=\! [v_{k,1},v_{k,2},...,v_{k,s}]^T \!=\! [\frac{1 \cdot C_m}{\prod_{j=1}^{k} s_j}, \frac{2 \cdot C_m}{\prod_{j=1}^{k} s_j},...,\frac{s_k \cdot C_m}{\prod_{j=1}^{k} s_j}]^T$.
The counting number $C_p$ of a local patch in mixture regression can be calculated as followed:

{\setlength\abovedisplayskip{-2mm}
\setlength\belowdisplayskip{0mm}
\begin{align}
C_p &= \sum_{k=1}^{K} {\bm{p}_{k}}^{T}  \bm{v}_{k} 
= \sum_{k=1}^{K} (\sum_{i=1}^{s_k} p_{k,i} \cdot \frac{i_k \cdot C_m}{\prod_{j=1}^{k} s_j}) \\
\label{eq_MRM} &= C_m \sum_{k=1}^{K} \sum_{i=1}^{s_k}  \frac{p_{k,i} \cdot i_k}{ \prod_{j=1}^{k} s_j }.
\end{align}
}

To illustrate the operation of MRM clearly, we take the regression with three mixtures ($K\!=\!3$) for example. For the first mixture, the length of each interval is $C_m/s_1$.  The interval is roughly divided, and the network learns a preliminary estimation of the degree of density, such as sparse, medium, or dense.
As the deeper feature in the network contains richer semantic information, we adopt the feature map of layer 5 for the first mixture.
For the second and third mixtures, the length of each interval is $C_m / (s_1 \!\times\! s_2)$ and $C_m / (s_1 \!\times\! s_2 \!\times\! s_3)$, respectively. Based on the fine estimation of the second and third mixtures, the network performs more accurate and detailed regression.
Since the shallower features in the network contain detailed texture information, we exploit the feature maps of layer 4 and layer 3 for the second and third mixtures of counting regression, respectively.

\subsection{Adaptive Soft Interval Module}{\label{Sec 3.4}}

In Sec \ref{Sec 3.3}, it is very inflexible to directly divide the regression interval into several non-overlapping intervals. The regression of value at hard-divided interval boundary will cause a significant error.
Therefore, we propose ASIM, which can shift and scale interval adaptively to make the regression process smooth. 

For shifting process, we add an extra interval shifting vector factor 
$\bm{\beta}_k \!=\! [{\beta}_{k,1}, {\beta}_{k,2},..., {\beta}_{k,s}]^T$ to represent interval shifting of the $i$-th interval of the $k$-th mixture, and the index of the $k$-th mixture $\overline{i}_k$ can be updated to:

{\setlength\abovedisplayskip{-2mm}
\setlength\belowdisplayskip{-2mm}
\begin{align}
\overline{i}_k = i_k + \beta_{k,i}.
\end{align}
}

For scaling process, similar to the shifting process, we add an additional interval scaling factor ${\gamma}$ to represent interval scaling of each mixture, and the interval number of the $k$-th mixture $\overline{s}_k$ can be updated to:

{\setlength\abovedisplayskip{-2mm}
\setlength\belowdisplayskip{-2mm}
\begin{align}
\overline{s}_k = s_k( 1 + \gamma_{k} ).
\end{align}
}


The network can get the output parameters $\{ \bm{p}_k, {\gamma}_k, \bm{\beta}_k \}$ for an input image.
Based on Eq. (\ref{eq_MRM}) and the given parameters $C_m$ and $s_k$, we can update the mixture regression result $C_p$ to:

{\setlength\abovedisplayskip{-2mm}
	\setlength\belowdisplayskip{0mm}
\begin{align}
C_p &= C_m \sum_{k=1}^{K} \sum_{i=1}^{s_k}  \frac{ p_{k,i} \cdot \overline{i}_k }{ \prod_{j=1}^{k} \overline{s}_j } \\
\label{regression}	&= C_m\sum_{k=1}^{K} \sum_{i=1}^{s_k}  \frac{ p_{k,i} \cdot {(i_k + \beta_{k,i})} }{ \prod_{j=1}^{k} [s_j(1+\gamma_k)] } .
\end{align}
}

Now, we detail the specific implementation of MRM and ASIM.
As shown in Fig. \ref{fig3}, for the feature maps from SAM, we downsample them to 
size $\frac{M}{w} \times \frac{N}{h}$ 
by following a two-stream model ($1 \times 1$ convolution and avg pooling, $1 \times 1$ convolution and max pooling) and channel-wise concatenation operation.
In this way, we can get the fused feature map from the two-stream model to avoid excessive information loss caused via down-sampling.
With linear mapping via $1 \times 1$ convolution and different activation functions (ReLU, Tanh and Sigmoid), we get regression factors $\{ \bm{p}_k, {\gamma}_k, \bm{\beta}_k \}$ , respectively.
We should note that, $\{ \bm{p}_k, {\gamma}_k, \bm{\beta}_k \}$  are the output of MRM and ASIM modules, only related to the input image. 
LCM is calculated according to parameters $\{\bm{p}_k, \gamma_k, \bm{\beta}_k \}$ and point-wise operation in Eq. (\ref{regression}).
Crowd number can be calculated via global summation over the LCM.
The entire network can be trained end-to-end.
The target of network optimization is $L_1$ distance between the estimated LCM
($ \mathit{LCM^{es}} $) and the ground-truth LCM ($ \mathit{LCM^{gt}} $):

{\setlength\abovedisplayskip{-2mm}
	\setlength\belowdisplayskip{-2mm}
\begin{align}
\label{loss}	\mathit{Loss} = \left\| \mathit{LCM^{es}} - \mathit{LCM^{gt}} \right\|_1.
\end{align}
}

\section{Experiments}{\label{Sec 4}}

In this section, we first introduce four public challenging datasets and the essential implementation details in our experiments. After that, we compare our method with state-of-the-art methods. Finally, we conduct extensive ablation studies to prove the effectiveness of each component of our method. 

\subsection{Datasets} 

We evaluate our method on four publicly available crowd counting benchmark datasets:
ShanghaiTech \cite{MCNN_2016_CVPR} Part A and Part B, UCF-QNRF \cite{Idrees_2018_ECCV} and
UCF-CC-50 \cite{Idrees_2013_CVPR}.
These datasets are introduced as follows.

\textbf{ShanghaiTech.} 
The ShanghaiTech dataset \cite{MCNN_2016_CVPR} is consisted of two parts: Part A and Part B, with a total of 330,165 annotated heads.
Part A is collected from the Internet and represents highly congested scenes, where 300 images are used for training and 182 images for testing.
Part B is collected from shopping street surveillance camera and represents relatively sparse scenes, where 400 images are used for training and 316 images for testing.

\textbf{UCF-QNRF.}
The UCF-QNRF dataset \cite{Idrees_2018_ECCV} is a large crowd counting dataset with 1535 high resolution images and 1.25 million annotated heads, where 1201 images are used for training and 334 images for testing. It contains extremely dense scenes where the maximum crowd count of an image can reach 12865.
We resize the long side of each image within 1920 pixels to reduce cache occupancy, due to the large resolution of images in the dataset.

\textbf{UCF-CC-50.}
The UCF-CC-50 dataset \cite{Idrees_2013_CVPR} is an extremely challenging dataset, containing 50 annotated images of complicated scenes collected from the Internet.
In addition to different resolutions, aspect ratios and perspective distortions, 
this dataset also has great variants of crowd numbers, varying from 94 to 4543.

\subsection{Implementation Details}

\textbf{Evaluation Metrics.}
We adopt mean absolute error (MAE) and mean squared error (MSE) as metrics to evaluate the accuracy of crowd counting estimation, which are defined as:

{\setlength\abovedisplayskip{-2mm}
	\setlength\belowdisplayskip{2mm}
\begin{align}
\mathit{MAE} = \frac{1}{N} \sum_{i=1}^{N} | C_i^{\ es}-C_i^{\ gt} |, \quad
\mathit{MSE} = \sqrt{ \frac{1}{N} \sum_{i=1}^{N} { ( C_i^{\ es}-C_i^{\ gt} )}^2 } ,
\end{align}
}where $N$ is the total number of testing images, $C_i^{\ es}$ (\emph{resp.} $C_i^{\ gt}$) is
the estimated (\emph{resp.} ground-truth) count of the $i$-th image, which can be calculated by summing the estimated (\emph{resp.} ground-truth) LCM of the $i$-th image.

\textbf{Data Augmentation.}
In order to ensure our network can be sufficiently trained and keep good generalization, we randomly crop an area of $m \times m$ pixels from the original image for training.
For the ShanghaiTech Part B and UCF-QNRF datasets, $m$ is set to 512.
For the ShanghaiTech Part A and UCF-CC-50 datasets, $m$ is set to 384. Random mirroring is also performed during training.

\textbf{Training Details.}
Our method is implemented with PyTorch.
All experiments are carried out on a server with an Intel Xeon 16-core CPU (3.5GHz), 64GB RAM and a single Titan Xp GPU.
The backbone of network is directly adopted from convolutional layers of VGG16 \cite{simonyan2014very} pretrained on ImageNet, and the other convolutional layers employ random Gaussian initialization with a standard deviation of 0.01.
The learning rate is initially set to $1e^{-5}$. The training epoch is set to 400 and the batch size is set to 1. We train our networks with Adam optimization \cite{kingma2014adam}  by minimizing the loss function Eq. (\ref{loss}).

\begin{table}[!t]
	\begin{center}
		\caption{Comparisons with state-of-the-art methods on ShanghaiTech Part A and Part B datasets}
		\label{tab_SH}
		\begin{tabular}{p{4cm}|p{1.4cm}<{\centering} p{1.4cm}<{\centering}
				|p{1.4cm}<{\centering} p{1.4cm}<{\centering}}
			\hline
			\quad Dataset & \multicolumn{2}{c|}{Part A} & \multicolumn{2}{c}{Part B} \\ \hline
			\quad Method  & MAE & MSE & MAE & MSE \\ \hline
			\quad MCNN \cite{MCNN_2016_CVPR} & 110.2 & 173.2 & 26.4 & 41.3 \\ 
			\quad Switch-CNN \cite{SwitchCNN_2017_CVPR} & 90.4 & 135.0 & 21.6 & 33.4 \\ 
			\quad CP-CNN \cite{CPCNN_2017_ICCV} & 73.6 & 106.4 & 20.1 & 30.1 \\  
			\quad CSRNet \cite{CSRNet_2018_CVPR} &  68.2 & 115.0 & 10.6 & 16.0 \\ 
			\quad SANet \cite{SANet_2018_ECCV} & 67.0 & 104.5 & 8.4 & 13.6 \\
			\quad PACNN \cite{shi2019revisiting} & 62.4 & 102.0 & 7.6 & 11.8 \\
			\quad SFCN \cite{wang2019learning} & 64.8 & 107.5 & 7.6 & 13.0 \\ 
			\quad Encoder-Decoder \cite{jiang_2019_CVPR} & 64.2 & 109.1 & 8.2 & 12.8 \\ 
			\quad Bayesian Loss \cite{BL_2019_ICCV} & 62.8 & 101.8 & 7.7 & 12.7 \\
			\quad RANet \cite{zhang2019relational}  & 59.4 & 102.0 & 7.9 & 12.9 \\  
			\quad PaDNet \cite{tian2019padnet} & \textbf{59.2} & \textbf{98.1} & 8.1 & 12.2 \\ \hline
			\quad Ours & 61.59 & 98.36 & \textbf{7.02} & \textbf{11.00} \\ \hline
		\end{tabular}
	\end{center}
\end{table}

\begin{table}[!t]
	\begin{center}
		\caption{Comparisons with state-of-the-art methods on UCF-QNRF and  UCF-CC-50 datasets}
		\label{tab_QNRF}
		\begin{tabular}{p{4cm}|p{1.4cm}<{\centering} p{1.4cm}<{\centering}
				|p{1.4cm}<{\centering} p{1.4cm}<{\centering}}
			\hline
			\quad Dataset & \multicolumn{2}{c|}{UCF-QNRF} & \multicolumn{2}{c}{UCF-CC-50} \\ \hline
			\quad Method  & MAE & MSE & MAE & MSE   \\ \hline
			\quad MCNN \cite{MCNN_2016_CVPR} & 277 & 426  &  377.6 & 509.1 \\ 
			\quad Switch-CNN \cite{SwitchCNN_2017_CVPR} & 228 & 445 & 318.1 & 439.2  \\
			\quad Composition Loss \cite{Idrees_2018_ECCV} & 132 & 191 & -- & -- \\
			\quad Encoder-Decoder \cite{jiang_2019_CVPR}  & 113  & 188 & 249.4 & 354.5  \\
			\quad RANet \cite{zhang2019relational} 	& 111	& 190 	& 239.8		& 319.4 \\  
			\quad S-DCNet \cite{S-DCNet_2019_ICCV} & 104.4 & 176.1 & 204.2 & 301.3  \\
			\quad SFCN \cite{wang2019learning} & 102.0 & 171.4 & 214.2 & 318.2  \\
			\quad DSSINet	\cite{liu2019crowd} & 99.1 & 159.2 & 216.9	& 302.4 \\
			\quad MBTTBF \cite{sindagi2019multi} & 97.5 & 165.2 & 233.1 &	300.9 \\
			\quad PaDNet \cite{tian2019padnet} & 96.5 & 170.2 & 185.8 & 278.3  \\
			\quad Bayesian Loss \cite{BL_2019_ICCV} & 88.7 & 154.8 & 229.3 & 308.2    \\ \hline
			\quad Ours & \textbf{86.6} & \textbf{152.2} & \textbf{184.0} & \textbf{265.8}  \\ \hline
		\end{tabular}
	\end{center}
\end{table}

\subsection{Comparisons with State of the Art}

The proposed method exhibits outstanding performance on all the benchmarks.
The quantitative comparisons with  state-of-the-art methods on four datasets are presented in Table \ref{tab_SH} and Table \ref{tab_QNRF}.
In addition, we also tell the visual comparisons in Fig. \ref{fig4}.

\textbf{ShanghaiTech.} 
We compare the proposed method with multiple classic methods on ShanghaiTech Part A
\& Part B dataset and it has significant performance improvement.
On Part A, our method
improves $ 9.69\% $ in MAE and  $ 14.47\% $ in MSE compared with CSRNet,
improves $ 8.07\% $ in MAE and  $ 5.42\% $  in MSE compared with SANet. 
On Part B, our method 
improves $ 33.77\% $ in MAE and $ 31.25\% $ in MSE compared with CSRNet, 
improves $ 16.43 \%$ in MAE and $ 19.12\% $ in MSE compared with SANet.

\textbf{UCF-QNRF.}
We then compare the proposed method with other related methods on the UCF-QNRF dataset. To the best of our knowledge, UCF-QNRF is currently the largest and most widely distributed crowd counting dataset.
Although the maximum sides of images are limited to $1920$ pixels, the proposed method achieves significant improvement over state-of-the-art methods.
For example, Bayesian Loss \cite{BL_2019_ICCV} achieves $88.7$ in MAE and $154.8$ in MSE, which currently maintains the highest accuracy on this dataset, while our method improves $2.37\%$ in MAE and $1.68\%$ in MSE, respectively.

\textbf{UCF-CC-50.}
We also conduct experiments on the UCF-CC-50 dataset. The crowd numbers in images vary from 96 to 4633, bringing a great challenging for crowd counting.
We follow the 5-fold cross validation as \cite{Idrees_2013_CVPR} to evaluate our method.
With a small amount of training images, our network can still converge well in this dataset.
Compared with the latest method Bayesian Loss \cite{BL_2019_ICCV}, our method improves $19.76\%$ in MAE and $18.18\%$ in MSE and achieves the state-of-the-art performance.

\subsection{Ablation Studies}

In this section, we perform ablation studies on ShanghaiTech dataset and demonstrate the roles of several modules in our approach.

\begin{figure}[!t]
	\centering
	\scalebox{0.43} {\includegraphics{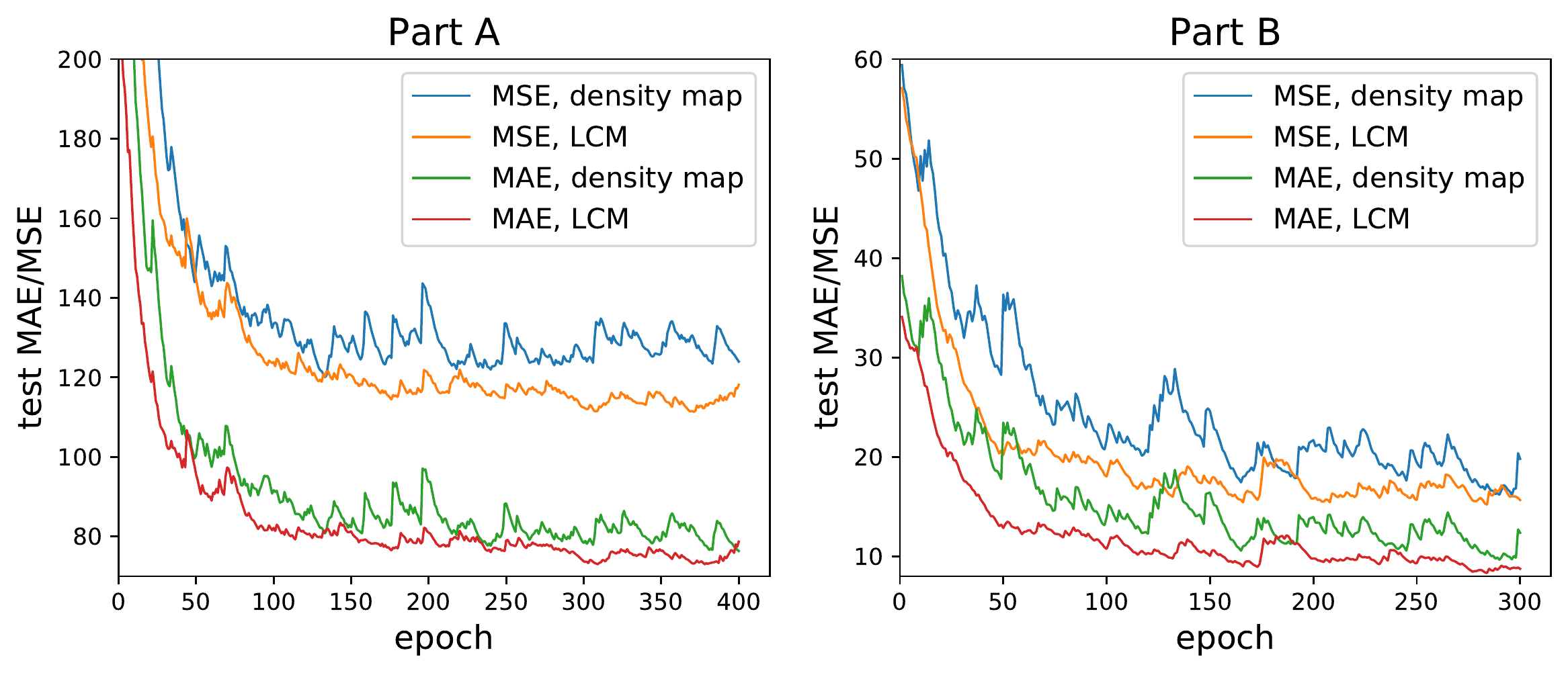}}
	\setlength{\belowcaptionskip}{-1mm}
	\caption{
			The curves of testing loss for different regression targets LCM and density map.
			LCM has lower error and smoother convergence curves on both MAE and MSE than density map
	}
	\label{fig_loss}
\end{figure}

\textbf{Effect of Regression Target.}
We analyze the effects of different regression targets firstly.
As shown in Table \ref{tab_LCM}, the LCM we introduced has better performance than the density map, with $4.74\%$ boost in MAE and $4.06\%$ boost in MSE on Part A, $8.47\%$ boost in MAE and $6.18\%$ boost in MSE on Part B. 
As shown in Fig. \ref{fig_loss}, LCM has more stable and lower MAE \& MSE testing curves.
It indicates that LCM alleviates the inconsistency problem between the training target and the evaluation criteria to bring performance improvement.
Both of them adopt VGG16 as the backbone networks without other modules.

\textbf{Effect of Each Module.}
To validate the effectiveness of several modules, we train our model with four different combinations:
1) VGG16+LCM (Baseline); 2) MRM; 3) MRM+ASIM; 4) MRM+ASIM+SAM.
As shown in Table \ref{tab_module},
MRM improves the MAE from $69.52$ to $65.24$ on Part A and from $8.96$ to $7.79$ on Part B, compared with our baseline direct LCM regression.
With ASIM, it improves the MAE from $65.24$ to $63.85$ on Part A and from $7.79$ to $7.56$ on Part B.
With SAM, it improves the MAE from $63.85$ to $61.59$ on Part A and from $7.56$ to $7.02$ on Part B, respectively.
The combination of MRM+ASIM+SAM achieves the best performance, 
$61.59$ in MAE and $98.36$ in MSE on Part A,
$7.02$ in MAE and $11.00$ in MSE on Part B.

\begin{table}[!t]
	\begin{center}
		\caption{An quantitative comparison with two different targets on testing datasets between LCM and  density map}
		\label{tab_LCM}
		\centering
		\begin{tabular}{p{3.5cm}<{\centering}|p{1.4cm}<{\centering} p{1.4cm}<{\centering}
				|p{1.4cm}<{\centering} p{1.4cm}<{\centering}}
			\hline
			& \multicolumn{2}{c|}{Part A} & \multicolumn{2}{c}{Part B} \\ \hline
			Target  & MAE & MSE & MAE & MSE \\ \hline			
			density map  &   72.98  &  114.89   &  9.79   &  14.40 \\
			local counting map &  \textbf{69.52}  &  \textbf{110.23}  &  \textbf{8.96}  & \textbf{13.51} \\ \hline
		\end{tabular}
	\end{center}
\end{table}

\begin{table}[t]
	\begin{center}
		\caption{Ablation study on different combinations of modules including MRM, ASIM and SAM in the regression framework}
		\label{tab_module}
		\begin{tabular}{p{3.5cm}<{\centering}|p{1.4cm}<{\centering} p{1.4cm}<{\centering}
				|p{1.4cm}<{\centering} p{1.4cm}<{\centering}}
			\hline
			& \multicolumn{2}{c|}{Part A} & \multicolumn{2}{c}{Part B} \\
			\hline
			Module  & MAE & MSE & MAE & MSE \\
			\hline
			LCM         	 &  	69.52    &   110.23    &      8.96      &  13.51    \\
			MRM   					 &  65.24	  &    104.81   &   7.79     &   12.55    \\ 
			MRM+ASIM   	    &   63.85     &    102.48   &   7.56    &   11.98  \\ 
			MRM+ASIM+SAM   &   \textbf{61.59}  &  \textbf{98.36}  &   \textbf{7.02}     &  \textbf{11.00}        \\ 
			\hline
		\end{tabular}
	\end{center}
\end{table}

\begin{table}[!t]
	\begin{center}
		\caption{The effects of different local patch sizes with MRM module}
		\label{tab_size}
		\begin{tabular}{p{2.5cm}<{\centering}|p{1.4cm}<{\centering} p{1.4cm}<{\centering}
				|p{1.4cm}<{\centering} p{1.4cm}<{\centering}}
			\hline
			& \multicolumn{2}{c|}{Part A} & \multicolumn{2}{c}{Part B} \\ \hline
			Size  & MAE & MSE & MAE & MSE \\ \hline
			$16 \times 16$  		&  70.45	   & 114.12    &  9.41       & 13.93       \\ 
			$32 \times 32$ 	   &  69.28	  & 109.24   &  8.68        & 13.44       \\ 
			$64 \times 64$   	  & \textbf{65.24}  &  \textbf{104.81}  &   \textbf{7.79}   &  \textbf{12.55} \\
			$128 \times 128$   &   67.73     &  105.15  &    7.93      &  12.78        \\ 
			\hline
		\end{tabular}
	\end{center}
\end{table}

\textbf{Effect of Local Patch Size.}
We analyze the effects of different local patch sizes on regression results with MRM.
As shown in Table \ref{tab_size}, the performance gradually improves with local patch size increasing and it slightly drops until $128\!\times\!128$ patch size. Our method gets the best performance with $64\!\times\!64$ patch size on Part A and Part B. 
When the local patch size is too small, the heads information that local patch can represent is too limited, and it is difficult to map weak features to the counting value.
When the local patch size is $ 1\!\times\!1$, the regression target changes from LCM to the density map.
When the local patch size is too large, the counting regression range will also expand, making it difficult to perform fine and accurate estimation.

\begin{figure}[!t]
	\centering
	\scalebox{0.40} {\includegraphics{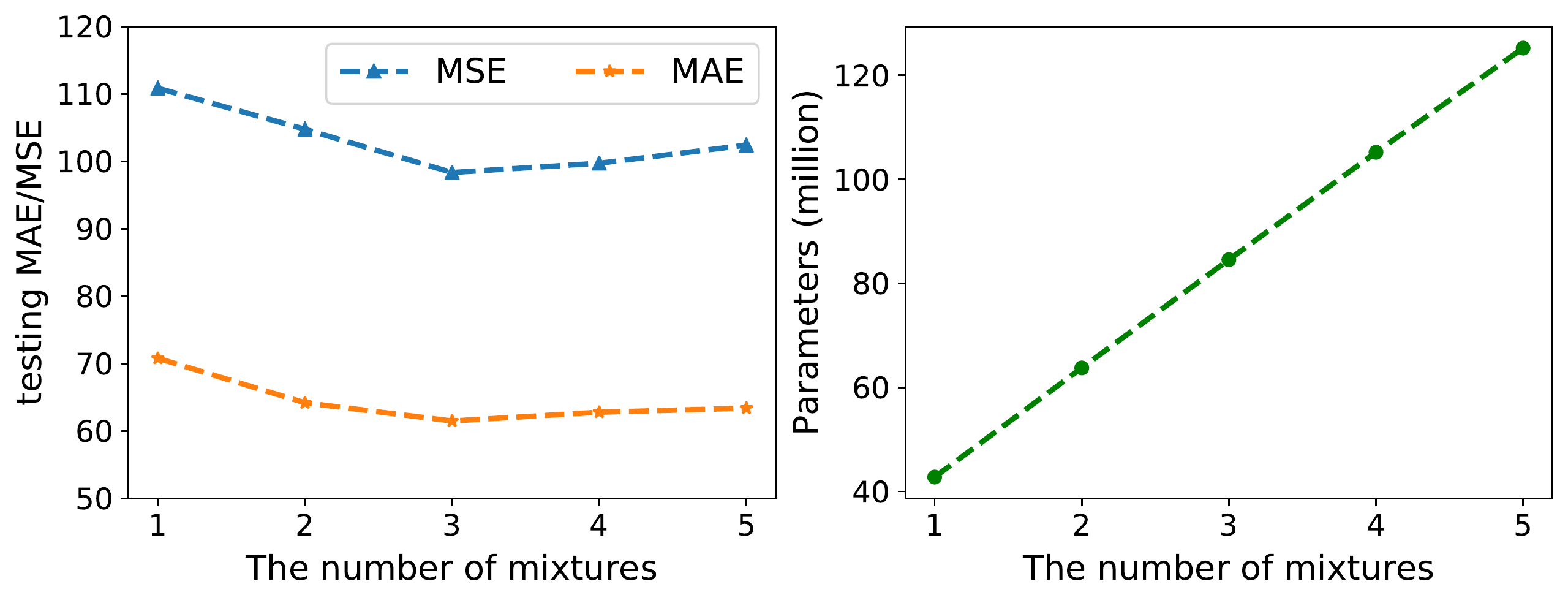}}
	\caption{		
		The left figure shows the MAE \& MSE performance of different mixtures number on ShanghaiTech Part A. The right one shows the relationship between the number of mixtures and the parameters of mixtures
	}
	\label{fig6}
\end{figure}

\textbf{Effect of Mixtures Number \it{K}.}
We measure the performance of adaptive mixture regression network with different mixture numbers $K$.
As shown in Fig. \ref{fig6}, the testing error firstly drops and then slightly arises with
the increasing number of $K$.
On the one hand, smaller $K$ (e.g., $K = 1$) means single division and it will involve a coarse regression on the local patch.
On the other hand, larger $K$ (e.g., $K = 5$) means multiple divisions.
It's obviously unreasonable when we divide each interval via undersize steps, such as $0.1$ and $0.01$.  
Besides, we analyze the relationship between the number of mixtures and model parameters.
As shown in Fig. \ref{fig6}, the increase of mixture number will also boost model parameters and calculation.
To achieve a proper balance between the accuracy and computational complexity, we take $K = 3$ as the mixtures number in experiments.

\section{Conclusion}{\label{Sec 5}}

In this paper, we introduce a new learning target named local counting map, and show its feasibility and advantages in local counting regression.
Meanwhile, we propose an adaptive mixture regression framework in a coarse-to-fine manner.
It reports marked improvements in counting accuracy and the stability of the training phase, and achieves the start-of-the-art performances on several authoritative datasets.
In the future, we will explore better ways of extracting context and multi-scale information from different convolutional layers.
Additionally, we will explore other forms of local area supervised learning approaches to further improve crowd counting performance.

\clearpage

\begin{figure}[!t]
	\centering
	\scalebox{0.2} {\includegraphics{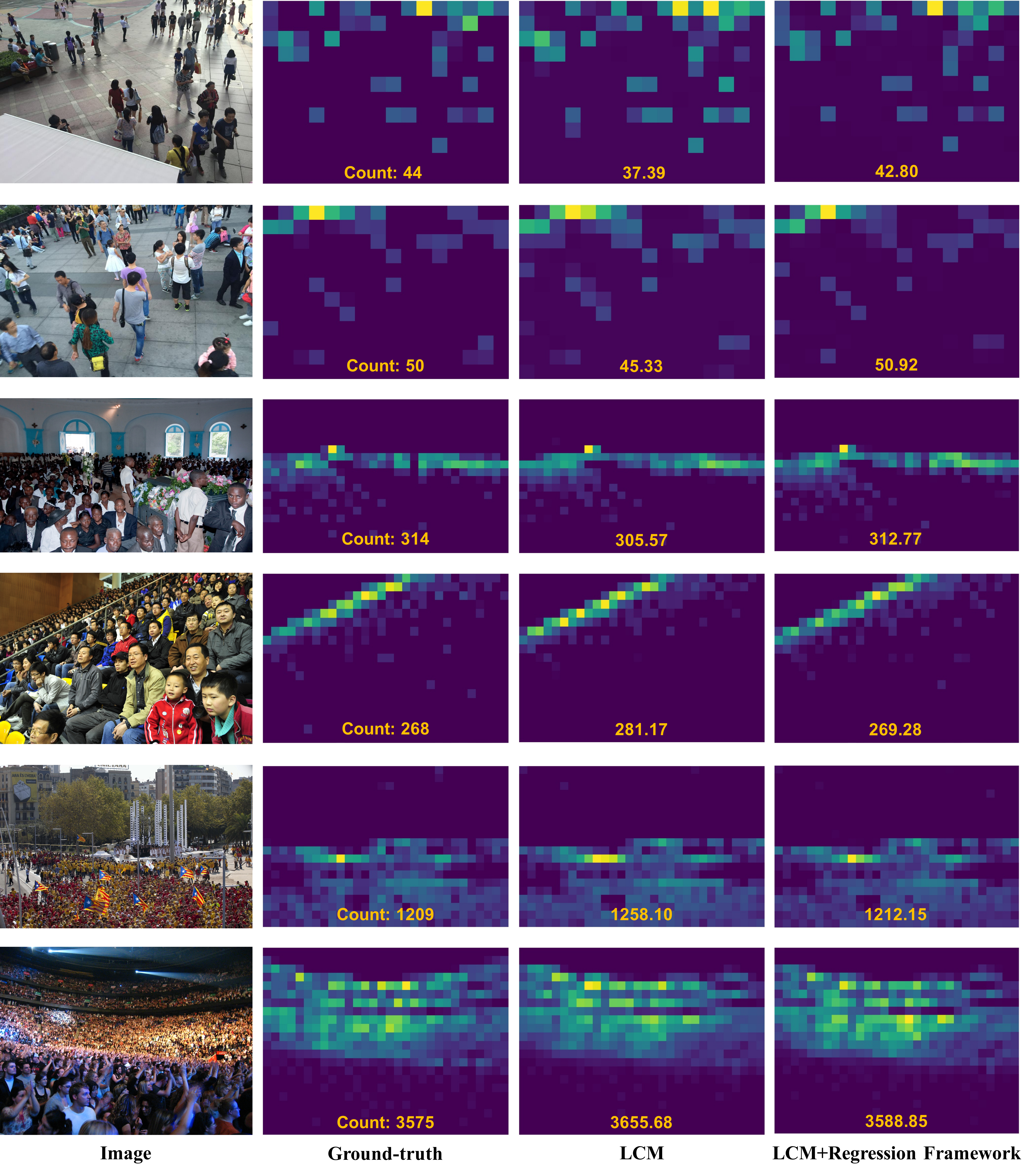}}
	\caption{
		From top to bottom, we exhibit several example images with different densities from sparse, medium to dense. The second column displays ground-truth local counting maps,
		the third column displays estimated LCM generated with the baseline model (VGG16+LCM),
		and the fourth column displays estimated LCM generated with our proposed regression framework.
		The brighter patch represents the larger counting value
	}
	\label{fig4}
\end{figure}



\clearpage
%
%

\bibliographystyle{splncs04}
\bibliography{egbib}
\end{document}